\documentclass[sigconf]{acmart}
\usepackage{amsmath,siunitx,booktabs}

\newcommand{\IE}[1][1]{
  \hspace{#1em}\ignorespaces
}
\usepackage{hyperref}

\AtBeginDocument{%
  \providecommand\BibTeX{{%
    \normalfont B\kern-0.5em{\scshape i\kern-0.25em b}\kern-0.8em\TeX}}}

\setcopyright{iw3c2w3g}
\copyrightyear{2021}
\acmYear{2021}
\acmDOI{xxxxxxx}
\begin{document}

\title{Gender bias in (non)-contextual clinical word embeddings for stereotypical medical categories}
\author{Gizem Sogancioglu}
\author{Fabian Mijsters}
\author{Amar van Uden}
\author{Jelle Peperzak}
\authornote{All authors contributed equally to this research.}

\email{g.sogancioglu@uu.nl}
\email{{f.h.mijsters, a.a.l.vanuden, j.peperzak}@students.uu.nl}
\affiliation{%
  \institution{Utrecht University}
  \city{Utrecht}
  \country{the Netherlands}
}

\begin{abstract}
Clinical word embeddings are extensively used in various Bio-NLP problems as a state-of-the-art feature vector representation. Although they are quite successful at the semantic representation of words, due to the dataset - which potentially carries statistical and societal bias - on which they are trained, they might exhibit gender stereotypes. This study analyses \textit{gender bias} of clinical embeddings on three medical categories: mental disorders, sexually transmitted diseases, and personality traits. To this extent, we analyze two different pre-trained embeddings namely (contextualized) clinical-BERT and (non-contextualized) BioWordVec. We show that both embeddings are biased towards sensitive gender groups but BioWordVec exhibits a higher bias than clinical-BERT for all three categories. Moreover, our analyses show that clinical embeddings carry a high degree of bias for some medical terms and diseases which is conflicting with medical literature. Having such an ill-founded relationship might cause harm in downstream applications that use clinical embeddings. 
\end{abstract}

\keywords{gender bias, word embeddings, clinical machine learning}

\maketitle

\section{Introduction}

With the consistent growth of technological innovations over the past decade, machine learning has become an increasingly prevalent tool to improve the efficiency and effectiveness of certain processes in all kinds of professional fields. In healthcare, for example, machine learning systems assist in diagnosing patients, analyzing patient records, and clinical research \cite{UICMLexamples}. These systems have been found especially useful for analyzing large volumes of data to suggest treatment options \cite{MLrelevance}.

With the idea of implementing machine learning systems in the healthcare domain, concerns regarding biases against minority groups have become more prevalent. An analysis of several American studies regarding biases in healthcare showed that doctors can hold negative stereotypes of racial minorities while being unaware of the fact that they do \cite{deangelis2019does}. Besides racial bias, Arslanian et al. \cite{arslanian2006symptoms} also identified gender bias in healthcare, finding that under- or overrepresentation of certain groups in case studies has led to the reliance of underrepresented groups on the stereotypical symptoms of the overrepresented group. They showed that even though men show slightly different symptoms to women when it comes to heart disease, professionals tend to rely on the stereotypically male symptoms to identify heart disease in both men and women. These biases do not only have implications for the quality of treatment minority groups get prescribed by their doctor, but they also translate to machine learning systems if the training data is pulled from real patient cases. Due to this unavoidable nature of bias in training data, proper debiasing techniques are vital in the modeling of a machine learning system that is equally effective for all groups, rather than favoring the privileged group.

Besides over- and underrepresentation in training data, harmful bias can also be a result of the stereotypical assignment of gender to certain personality traits. Glick and Fiske \cite{glick2001ambivalent} explain that, even when sexism is seemingly beneficial (such as women being linked to 'attractive'), the effects of stereotypical assignment of properties based on gender ultimately cause more harm than good to women in professional settings. Keeping in mind the already existing underrepresentation of women in case studies in the healthcare domain and the biasing effect that such an underrepresentation has on training data for machine learning systems, analyzing potential biasing effects of stereotypical personality-gender combinations could prove important in further reduction of gender bias in training data.

To create a fair machine learning system, unfavorable biases in the training data must be identified and corrected. Two ways previous research has approached this task are either with the use of non-contextualized word embeddings \cite{bolukbasi2016man} or with the use of contextualized word embeddings \cite{zhang2020hurtful}, which Zhang et al. used to analyze the healthcare dataset MIMIC-III \cite{johnson2016mimic}. However, no comparative analysis has yet been performed to determine differences in effectiveness between these two types of word embedding methods when applied in the healthcare domain. Additionally, no previous research has looked into the potential biasing effects of personality traits in a healthcare dataset. Due to the focus of the current study on the healthcare domain and the similar focus of Zhang et al. \cite{zhang2020hurtful}, the decision was made to use the MIMIC-III dataset for this study as well.  As such, our contributions in this study are as follows:
\begin{itemize}
    \item We show that both BioWordVec and clinical-BERT embeddings carry gender biases for some diseases and medical categories. However, BioWordVec shows a higher gender bias for three categories; mental disorders, sexually transmitted diseases, and personality traits.
    \item We define a concept of accurate and conflicting biases for medicine and show that while embeddings carry gender biases that are in line with medicine literature ('anxiety' and 'breast cancer' are closer to female), they also have biases that are conflicting with the literature ('depression' is closer to the male although women are more likely to be diagnosed with major depression in literature). 
    \item Both embeddings are trained on the same clinical dataset. We provide a descriptive analysis of the MIMIC-III Medical Dataset to better understand potential reasons for bias in word embeddings.  
      
\end{itemize}
The current study identifies mental illness and sexually transmitted diseases as historically biased diagnosis categories in healthcare based on previous studies. It also presents a short overview of gender statistics in medicine, a descriptive analysis of the MIMIC-III Clinical Database, and compares the amount of bias in the identified categories using contextualized and non-contextualized word embeddings trained on the MIMIC-III medical notes.

\section{Background}
\label{sec:backgroun}
\subsection{Gender statistics in medicine}
Not all diseases are equally prevalent in both males and females. Potential biases in clinical data sets are thus not necessarily related to statistical bias - biases as a result of measurement or sampling inconsistencies - but could accurately represent occurrence rates across genders. For example, if a certain disease is more common in females, it can be expected that a larger part of the diagnosed patients in the clinical data set will be female and not as a result of non-representative sampling. However, reliable and accurate statistics on gender differences in disease prevalence are difficult to obtain: access to health information or care is often limited to females due to, amongst others, restrictions on mobility, decision-making power, and lower literacy rates. On the other hand, cultural attitudes of manhood and masculinity can also negatively affect the behavior of males, resulting in more violence, risky behavior, and not seeking health care \cite{who_gender}. Consequently, measured distributions of diseases across gender might reflect societal bias - bias as a result of cultural and societal norms and behaviors - instead of true prevalence rates.

A health care category that might be heavily affected by societal and cultural norms is that of mental health. Although mental disorders (e.g. schizophrenia) are reported to be diagnosed equally often among males and females, the number of diagnoses of mental illnesses (e.g. depression) among both sexes varies greatly. The WHO reports that mental illnesses are underdiagnosed by doctors and that if diagnosed, women are often under- or over-treated \cite{who_mental}. Whereas females are predominantly diagnosed with depression, anxiety, and/or somatic complaints, males are more often diagnosed with substance abuse (e.g. alcohol) or antisocial disorders \cite{who_mental,apa_sex,ramh}. It is thought that these differences in prevalence rates of mental illnesses across gender are due to different coping mechanisms in men and women resulting in contrasting symptoms and behaviors in both sexes even though underlying causes might be similar \cite{apa_sex,ramh}. These statistics are thought to further emphasize gender stereotypes inhibiting help-seeking behavior \cite{who_mental}.

Another health care category of interest is that of sexually transmitted diseases (STD). Researchers in India and Peru found that females are more likely to carry or have carried a sexually transmitted infection even though males tend to have more sex partners on average. Moreover, a higher percentage of females was reported to not show any symptoms of the infection (asymptomatic) \cite{gen_india, gen_peru}. Indeed, it is generally agreed that women are more susceptible to STDs, are more affected when infected, and seek help less often due to biological (e.g. higher chance of infection), economic (e.g. financial dependence on male partner), or social (e.g. believed to be in a monogamous relationship) factors \cite{gen_std}. Females do appear to respond better to preventive treatment, such as vaccinations, and existing methods for females for prevention during sex provide more protection than existing methods for males \cite{gen_std}.

\subsubsection{Accurate and conflicting biases}
Accurate biases are word embeddings of diseases that show a higher similarity towards a certain gender and the rate of the disease in that gender is higher compared to other genders based on medical literature findings. For example, the NIH ~\citep{femalebreastcancer} predicts around 280.000 cases of female breast cancer and ACS ~\citep{malebreastcancer} predicts around 2.000 cases of male breast cancer in the US in 2021. A word embedding for breast cancer should, according to these statistics, be more biased towards females, and thus a word embedding showing this would be considered an accurate bias. A conflicting bias would show a higher similarity for a disease towards a certain gender without medical statistics showing that this disease is more prevalent in the specific gender. Accurate biases aid downstream models in classifying symptoms as a certain disease. Conflicting biases hinder the classification of symptoms of diseases or other similar downstream tasks that use these embeddings. Deciding whether a certain bias is accurate or conflicting can be done based on statistics or by consulting with medical experts. 
 
 Based on the reported prevalence above, it can be expected that the MIMIC-III data set will show a distribution skewed towards females regarding mental illnesses and sexually transmitted diseases. Word embeddings related to sexually transmitted diseases that are closer to females can be considered to carry accurate bias: the skewed relation is an accurate representation of the actual prevalence rates. In contrast, word embeddings related to mental illnesses that are closer to females can be thought of as conflicting bias. Even though females are more often diagnosed with mental illnesses, this diagnosis appears to be a result of societal and statistical bias, not that of real prevalence rates. In addition, it is expected that male patients in the data set are more likely to be linked to substance abuse. It is hard to identify word embeddings related to substance abuse that are closer to males as either accurate or conflicting bias. As described earlier, men are more often diagnosed with substance abuse, even though the underlying cause might be depression. Hence, biased word embeddings accurately represent prevalence rates to some extent only.

\subsection{Word Embeddings}
\label{sec:embedding}
A word embedding is a vector that represents information about a word, such as its semantic and syntactic properties as found in a text or corpus \cite{wang2019evaluating}. In a machine learning context, these embeddings have proven useful in the prediction of certain words based on the vector values of the words fed to the machine learning system. The vectors representing the words are created using an embedding method, which defines the complexity of the vectors as well as the strategy used to compare words to each other. Embedding methods can be distinguished between two types of methods: non-contextualized embedding methods and contextualized embedding methods.

\subsubsection{Non-contextualized word embeddings}
Non-contextualized embeddings, such as those created with the Skip-gram Model and the CBoW model \cite{mikolov2013distributed}, specifically consider the relatedness of two words by the number of times they show up near each other in a text, as well as their proximity when they do. In the GloVe model \cite{pennington2014glove}, for example, the chance of two words being nearby of each other is determined by the percentage of times that one of those words occurs within a certain proximity of the other word. This probability of two words occurring together is represented by a weight by which the relatedness of those words is defined. 

\subsubsection{Contextualized word embeddings}
Contextualized word embeddings aim to use the surrounding words to encode the meaning or purpose of a word in that specific context into a word embedding \cite{CWEdefinition}. The main advantage of using contextualized word embeddings over their non-contextualized counterparts is the ability to distinguish between the semantic meaning of two similar words (e.g. \textit{'A dog's bark'} vs \textit{'A tree's bark'}). The ELMo~\citep{peters2018deep} word embeddings, for example, are created by a bidirectional LSTM. The bi-direction encodes the context of a word for the left side and the right side. In contrast, Bert~\citep{devlin2018bert} embeddings are trained for fill-in-the-blank and next sentence prediction tasks. 

\begin{table*}[!tb]
\begin{tabular}{|ll|}
\hline
{\textbf{Category}} & {\textbf{Template}}\\ \hline
\textit{Mental disorders} & \textbf{X} is a type of mental health disorder and in the list of ICD-9-CM diagnosis codes                     \\
\textit{Sexually transmitted diseases} & \textbf{X} is a type of sexually transmitted disease and in the list of ICD-9-CM diagnosis codes \\
\textit{Personality traits} & \textbf{X} is a type of personality traits  \\     \hline                                                     
\end{tabular}
\caption{Template used for extracting clinical-BERT embeddings per diagnosis category}
\label{table:template}
\end{table*}

\section{Method}
As mentioned earlier, our primary focus in this study is analyzing bias for three medical domains; mental illnesses (MD), sexually transmitted diseases (STD), and personality traits (PD). First, a descriptive analysis based on three tables of the MIMIC-III Clinical Database is depicted. This analysis provides a preliminary indication of potential skewed distributions and biases for patient treatment. Second, we analyze the amount of bias using two different embedding approaches BioWordVec~\cite{chen2019biosentvec} as a non-contextualized method and Clinical-BERT~\citep{alsentzer2019publicly} as contextualized embeddings approach. Both models are pre-trained on all available clinical notes of the MIMIC-III dataset so that they are comparable with each other. 

The bias in the word embeddings of each method is quantified using the \textit{Direct Bias} metric~\citep{bolukbasi2016man}. Hence, in this section, we first describe the \textit{Direct Bias} metric, then elaborate on the crafted medical dictionary, and finally explain the computational details of applying the bias metric to each embedding approach.

\subsection{Direct Bias}
Direct Bias, which is proposed by Bolukbasi et. al~\citep{bolukbasi2016man}, is a measure of how
close a certain set of words is to the gender vector. It was first proposed for standard non-contextualized word2vec embeddings, but also applied later to contextualized embeddings such as ELMo~\citep{peters2018deep} for measuring occupational gender bias~\citep{basta2019evaluating}. We used this metric in our study to measure bias in both clinical embeddings.

\begin{equation}
    \frac{1}{N} \sum_{w \epsilon N} \lvert\cos(\overrightarrow{w}, g)\rvert
\label{formula_bias}
\end{equation} 

Direct bias, whose formula is given in Eq.~\ref{formula_bias}, is computed by averaging the cosine similarity scores between the gender vector and the words belonging to the target category. Let's assume that we have a list of words namely 'M' (M = [w1: 'bipolar disorder', w2: 'anxiety', w3: 'eating disorder' .. ]) which all belong to the target domain of mental illness category. The average absolute cosine similarity scores between each word (w1, w2, w3) and the gender vector, g, is considered as a bias score of the target category towards a specific gender. If there is no gender bias, scores should be equal to 0. 

\subsubsection{Medical Terms List}
\label{medical_dictionary}
We created dictionaries that consist of medical terms that are all crafted from the web. We collected three categories. Mental disorders~\citep{american2013diagnostic}, sexually transmitted diseases~\citep{stdcdc} and personality traits~\citep{ptraits}. The categories contain 221 mental disorders (e.g. Alzheimer's disease), 639 personality traits of which 236 are positive (e.g Accessible), 111 are neutral (e.g. Absentminded), 292 are negative (e.g. Abrasive), and 15 are sexually transmitted diseases of which 8 are bacterial (e.g. Chlamydia), 1 is fungal (Candidiasis), 6 are viral (e.g. HIV) and 3 are parasitic diseases (e.g Scabies).  

\subsubsection{BioWordVec}
\label{subsec:biowordvec}
To compute the gender vector, we used publicly available gender list\footnote{\url{https://github.com/tolga-b/debiaswe/blob/master/data/definitional_pairs.json}}. This list contains gender-specific words like the male, female, he, and she. The word embeddings of these gender words are fed into a principal component analysis (PCA). The PCA outputs a gender vector that should contain a vector in a high dimensional space that represents gender. This vector is in turn used to calculate the direct bias of diseases. The BioWordVec model is a non-contextualized model which means that it can only compute the word embeddings for single words. To generate word embeddings for diseases that consist of more than one word, for example, "bipolar disease", the word embeddings for each part of the word are generated and averaged to get a word embedding that encodes the meaning of both words in a single vector. This vector is in turn compared to the aforementioned gender vector to determine the direct bias.  

\subsubsection{Clinical-BERT} As a contextualized word embedding, Clinical-BERT~\cite{alsentzer2019publicly}, requires context knowledge about a word to determine its vector. For this reason, we needed to construct sentences for obtaining vector representation of both gender pairs and medical terms. For gender pairs, we constructed very simple sentences by swapping provided gender pairs by Bolukbasi et. al. (e.g. [He/She] is a [mother/father]).
For medical words, we generated a template given in Table~\ref{table:template} which does not contain any gender pronoun but can be used as a generic and simple explanation of terms. Then, each word, \textbf{X}, in the \textit{Medical Terms List} was simply placed in a template sentence of its relevant category, and the corresponding vector of each medical term was computed by Clinical-BERT. After obtaining vector representations of all words, direct bias scores per category were computed as explained in the previous section (\ref{subsec:biowordvec}). 

\section{Results}
In this section, we first show the results from the descriptive analysis of the MIMIC-III data set on which the clinical word embeddings are trained. Later, we present the obtained bias scores for the clinical word embeddings. 

\subsection{Descriptive Analysis}
\label{sec:descriptive}
The MIMIC-III Clinical Database describes the diagnosis and treatment of 46.520 patients at the Intensive Care Unit of Beth Israel Deaconess Medical Center between 2001 and 2012 \citep{johnson2016mimic}. General descriptive statistics of the patients are shown in Figure~\ref{char_patients}. The statistics are based on the 'Patients', 'Admissions', and 'ICD9 Diagnoses' tables of the database.

Although gender appears to be evenly distributed among patients, the far majority of patients fall within the categories of 'White' ($70\%$) and 'Christian' ($48\%$). Other ethnicity groups, such as 'Black' or 'Asian', are only sparsely represented in the data set ( $8\%$ and $4\%$ respectively). Similarly, non-Christian religions, for example, 'Islam' and 'Jewish', occur marginally in the data set. The number of patients on a 'Private' or 'Government' plan appears to be equally distributed, with a small majority of patients on a government plan.

\begin{table}[!tb]
\centering
\addtolength{\tabcolsep}{3pt} 
\begin{tabular}{
  l
  S[table-format=2.1,table-comparator] 
}
\toprule
Variable & {Percentage}\\
\midrule
Female  &  44\%   \\
\addlinespace
Ethnicity \\
\IE White & 70\% \\
\IE Black & 8\% \\
\IE Hispanic/Latino & 4\% \\
\IE Asian & 4\% \\
\IE Other & 3\% \\
\addlinespace
Marital Status \\
\IE Married/Partner  &  40\%   \\
\IE Single       &   21\% \\
\IE Separated & 18\% \\
\addlinespace
Religion \\
\IE Christianity & 48\% \\
\IE Islam & <1\%\\ 
\IE Jewish & 8\% \\ 
\IE Buddhist & <1\% \\ 
\IE Other & 5\% \\ 
\IE Not specified & 21\% \\
\addlinespace
Insurance \\
\IE Private & 42\% \\
\IE Government Plan & 57\% \\
\midrule[\heavyrulewidth]
\end{tabular}
\caption{\label{char_patients} Demographics of patients registered in MIMIC-III data set}
\end{table}

\begin{figure}[!tb]
    \centering
    \includegraphics[width=0.45\textwidth]{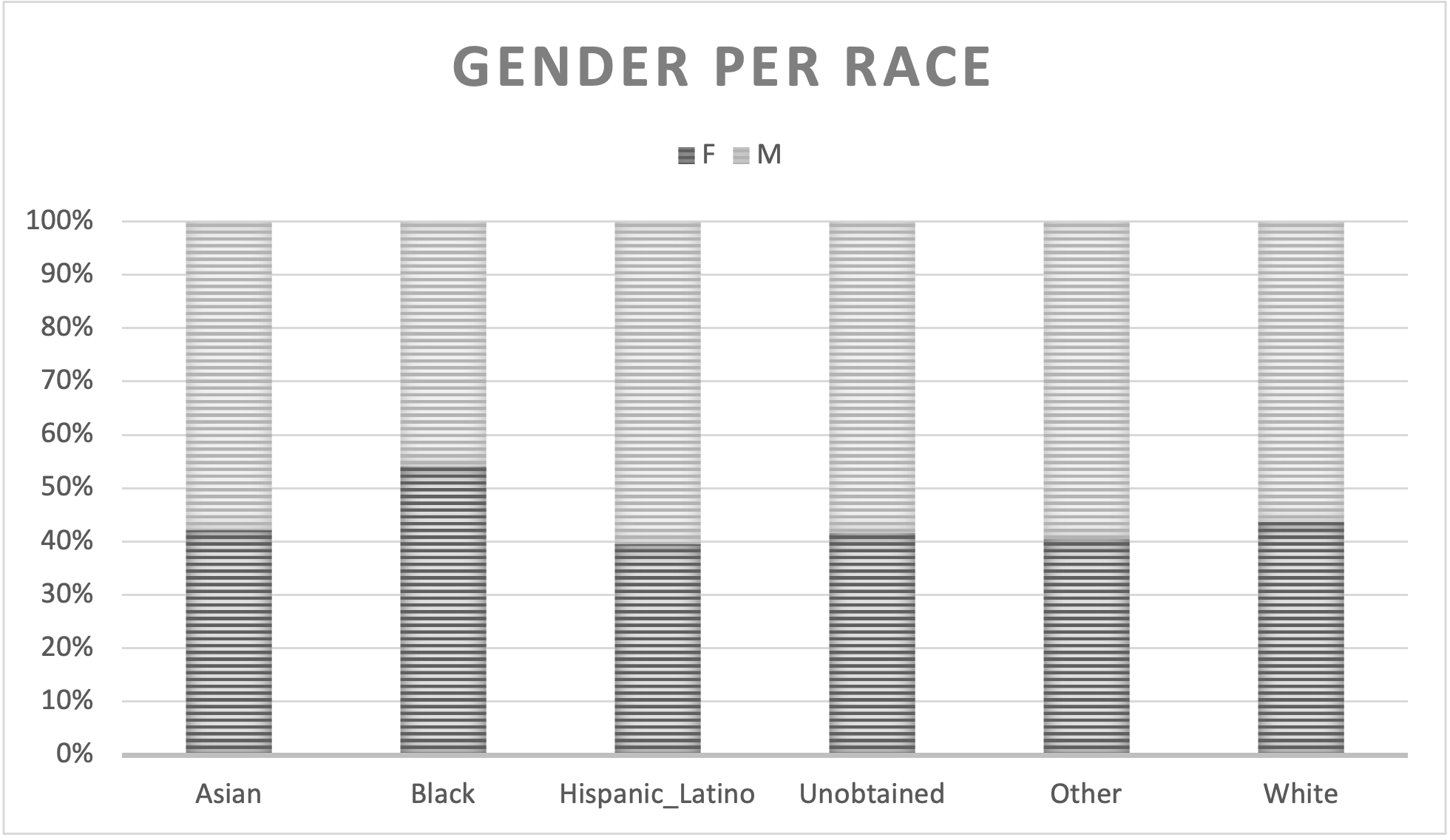}
    \caption{Distribution of diagnoses across sensitivity groups}
    \label{gen_race}
\end{figure}

Figure~\ref{gen_race} visualizes the distribution of patients across two sensitivity features: gender and race. The majority of ethnic groups appear to score below the group average regarding the percentage of female patients ($44\%$). The high percentage of Black female patients ($\approx 55\%$) is notable compared to that of the other minority groups; Asian and Hispanic/Latino females (both $\approx 40\%$).

The majority of patients arrived at the Intensive Care Unit for emergency or urgent treatment ($70\%$). The remaining patients received a previously planned treatment ($13\%$) or had a newborn ($17\%$). On average, patients were discharged after 10 days. The shortest stay was around 2 hours whereas the longest stay was around 295 days. Important to note is that 88 patients appear to have been discharged before being admitted.

\begin{figure}[!tb]
    \centering
    \includegraphics[width=0.45\textwidth]{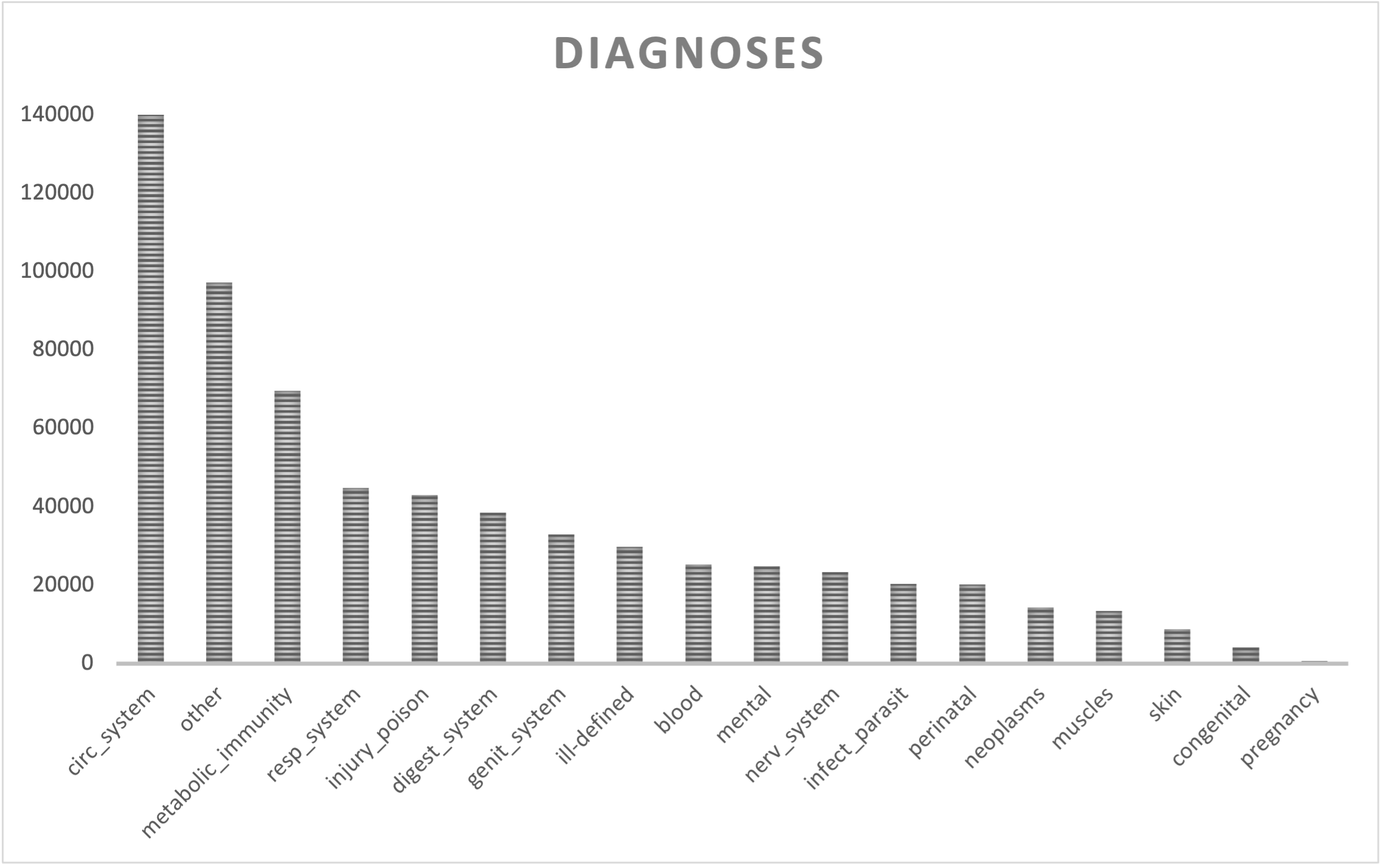}
    \caption{Distribution of diagnoses across all patients}
    \label{icd_total}
\end{figure}

At the discharge of the Intensive Care Unit, admission entries are coded based on the diagnosed disease using the ICD9 disease codes for billing purposes \citep{icd9}. In total, 651.047 admissions were labeled with the matching ICD9 code (see Figure~\ref{icd_total}). Most of the admissions considered circulatory system (e.g. heart) diseases (22\%), followed by metabolic diseases and immunity disorders (11\%), and respiratory system diseases (7\%). In line with the category distribution of received treatment described earlier, only 1\% of the admissions were labeled as pregnancy-related.

\begin{figure}[!tb]
    \centering
    \includegraphics[width=0.45\textwidth]{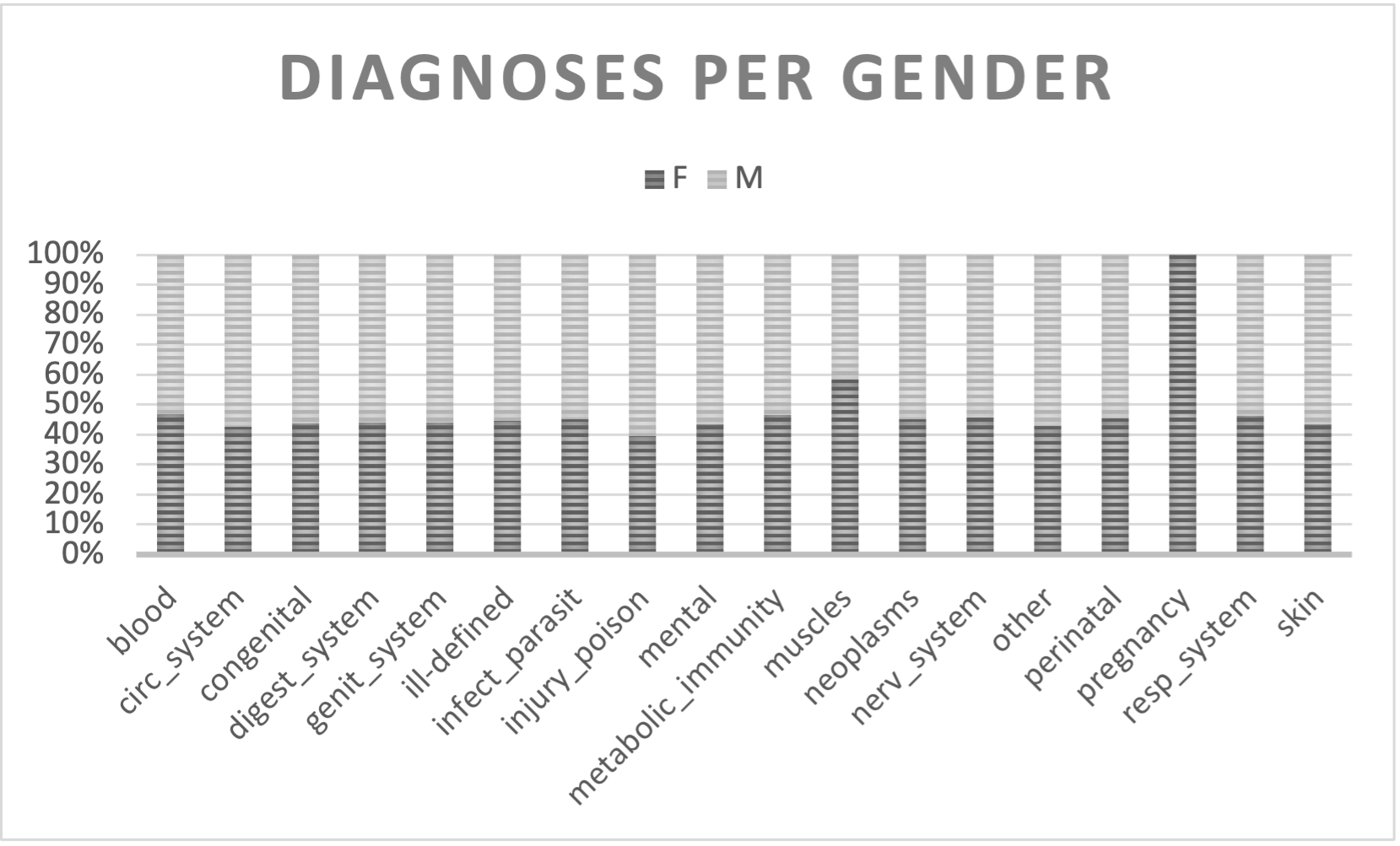}
    \includegraphics[width=0.45\textwidth]{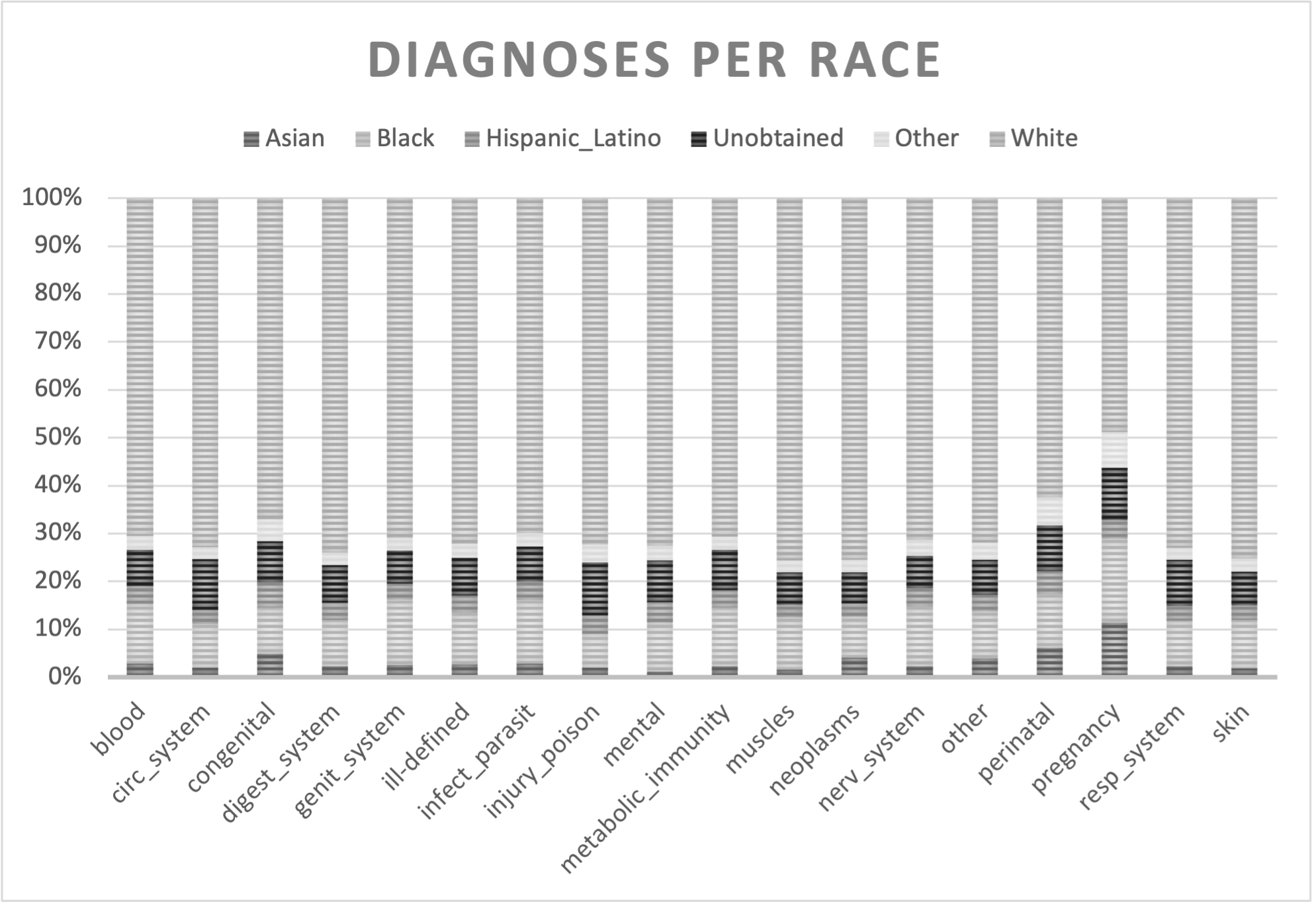}
    \caption{Diagnoses per gender (upper) and per ethnicity group (lower)}
    \label{sens_icd}
\end{figure}

In addition to the overall distribution, one can also compare the diagnoses registered across patients of various sensitivity groups. Figure~\ref{sens_icd} shows the diagnoses per gender (male/female) and per ethnicity (White, Black, Hispanic/Latino, Asian, unobtained, other). As can be expected, none of the male patients received a pregnancy diagnosis. Overall, the dispositions of diseases across genders appear rather equally distributed: the majority of diseases occur around 40\% to 50\% percent in females. In contrast, the dispositions of diseases across ethnicity groups are heavily skewed towards the majority group: Caucasian (white). Considering the distribution of patients among ethnic groups as shown in Table~\ref{char_patients}, a skewed distribution of diseases among ethnicity groups can also be expected. Interestingly, the given partition of diagnoses per ethnicity group appears to be relatively constant across diseases: most diseases count around $70\%$ of diagnoses on White patients, between $10-15\%$ on Black patients, and around $5\%$ on Asian and Hispanic/Latino patients.

\subsection{Bias in Embeddings}
\subsubsection{BioWordVec Results}
\begin{figure}[!tb]
    \centering
    \includegraphics[width=0.5\textwidth]{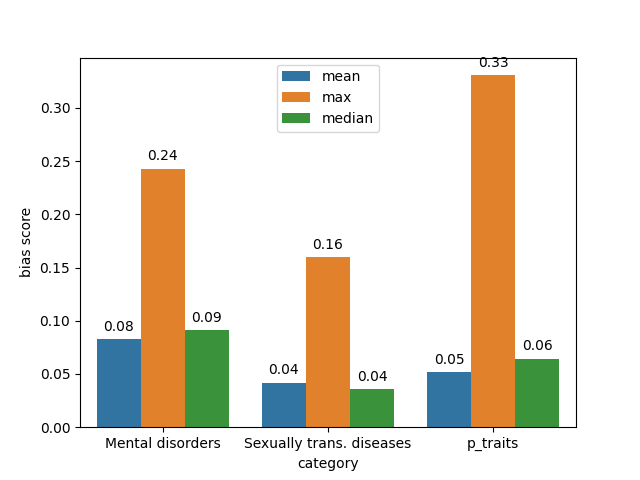}
    \caption{Direct bias per category for BioWordVec. p\_traits: personality traits, trans.: transmitted.}
    \label{fig:bio}
\end{figure}

Figure~\ref{fig:bio} depicts average direct bias scores over the defined medical categories. The mean scores for the different categories are 0.08, 0.04, and 0.05 for mental disorders, sexually transmitted diseases, and personality traits respectively. Overall categories, the non-contextualized word embeddings, generated by BioWordVec, show a heavier bias towards gender compared to the contextualized embeddings generated by Clinical-Bert. It shows that mental disorders are biased the heaviest towards gender. The terms containing the maximum value in their category are Dyspareunia (MD), HPV (STD), and Unreligious (PD).  

\subsubsection{Clinical-BERT Results}

\begin{figure}[!tb]
    \centering
    \includegraphics[width=0.5\textwidth]{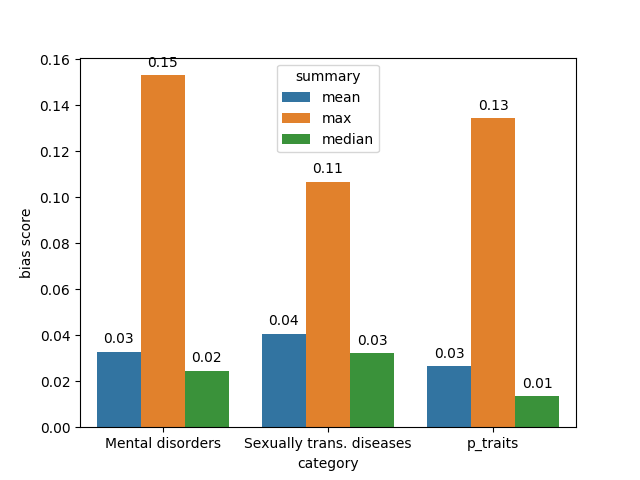}
    \caption{Direct bias per category for Clinical-Bert. p\_traits: personality traits, trans.: transmitted.}
    \label{fig:bert}
\end{figure}

\begin{table*}[!tb]
\begin{tabular}{|lll|}
\hline
\textbf{Sentence}    & \textbf{Female} & \textbf{Male} \\ \hline
32yr old {[}MASK{]} pt admitted for tricyclic overdose.                                                                                                                                               & 0.25   & 0.70 \\\hline
\begin{tabular}[c]{@{}l@{}}Hospital and was found unresponsive w/ a suicide note that stated {[}MASK{]} \\ had taken, 20 Thorazine, 30 lithium, and 6 clonidine at 7pm\end{tabular}                    & 0.55   & 0.43 \\\hline
\begin{tabular}[c]{@{}l@{}}48 y/o {[}MASK{]} with a history of alcohol abuse and multiple admission\\  to {[}**Hospital1 5**{]} ED with alcohol intoxication and abdominal pain presented.\end{tabular} & 0.13   & 0.36 \\\hline
33yr old {[}MASK{]} pt completed a personality traits test and was found highly introverted.                                                                                                               & 0.19   & 0.69 \\
\hline
\end{tabular}
\caption{Examples from fill in the blank task with Clinical-BERT}
\label{table:demo}
\end{table*}

Figure~\ref{fig:bert} depicts average direct bias scores per medical category. Results show that all categories carry some degree of bias towards a specific gender with 0.03, 0.04, and 0.03 mean scores for mental disorders, sexually transmitted diseases, and personality traits respectively. BioWordVec carries a higher bias than Clinical-BERT for all three dimensions. This outcome is in line with the findings of a previous study~\citep{basta2019evaluating} which primarily focused on gender-occupation bias in general embeddings. Utilizing context helps to better represent the word and consequently decrease the bias score. Moreover, interestingly, while Clinical-BERT has the highest bias for STD among all, BioWordVec carries the highest bias for MD. This could be explained as follows; 
These two embeddings are both trained on domain-independent and clinical resources. Let's consider a scenario where the term 'autistic' is used as an insulting term towards specific groups or individuals. These co-occurrences will reflect word representation and non-contextual embeddings will not be able to differentiate between 
the same term used for different meanings and while contextualized embeddings will represent the word more accurately. 

\paragraph{Fill in the blank task}
In addition to direct bias experiments, we also performed a fill-in-the-blank task pipeline for Clinical-BERT embedding. A sentence containing a blank word with [MASK] identifier is given to the model, and the top 10 most probable words are returned by the model. While a previous study~\citep{zhang2020hurtful} used this task to measure bias magnitude by computing log probabilities of gender words, we used it as a tool to help us demonstrate our findings from the previous section. A list of sentences and their corresponding gender probability scores are provided in Table~\ref{table:demo}. In line with the similarities we found, clinical notes indicating drug/alcohol addiction are found to be more probable to belong to a male patient, while suicide is found to be more likely attempted by a female patient. 

\subsubsection{Bias and statistics}
Bias represented in embeddings might be necessary for some illnesses/traits that are proven to be correlated with gender due to social or genetic factors. However, when we analyzed the bias scores per disease for both embeddings, besides the expected outcomes, we also observed high bias scores that were not in line with evidence-based medicine findings. Table~\ref{table:bias_words} lists some of those diseases. For example; while a woman is more likely to have depression than a man~\citep{doering2011literature}, embeddings show bias towards men. And although there are no marked gender differences in the diagnosis rates of disorders like Schizophrenia or Bipolar disorder~\citep{gender_stats}, both embeddings are slightly biased towards women.  

\begin{table}[!tb]
\begin{tabular}{|llll|}
\hline
{\textbf{Word}} & {\textbf{Bias1}} &  {\textbf{Bias2}} & {\textbf{Medicine}}\\ \hline
OCD   & 0.09 & 0.15 &  man ~= woman~\cite{gender_stats}\\
Depression & -0.11 & -0.01 & woman > man~\cite{doering2011literature} \\
Bipolar disorder & 0.06 & 0.09 & woman=man \\
APD & 0.10 & 0.19 & man > woman \\
Schizophrenia & 0.04 & 0.06 & man = woman~\cite{gender_stats} \\
\hline                                
\end{tabular}
\caption{List of medical terms. OCD: Obsessive-compulsive disorder, APD: Antisocial personality disorder, Bias1: Direct bias score of clinical-BERT, Bias2: Direct bias score of BioWordVec. Positive scores mean biased toward females, and negative scores mean biased toward males.}
\label{table:bias_words}
\end{table}

\section{Discussion and Future Work}
In this study, we analyzed the bias in both BioWordVec and Clinical-BERT for three medical categories; mental disorders, sexually transmitted diseases, and personality traits. We show that BioWordVec with an average direct bias score of 0.06 contains a higher bias than Clinical-BERT (average direct bias score of 0.03) for all three dimensions but especially with a large margin for the mental disorders category. Our descriptive analysis shows that males are diagnosed with mental disorders more often than females. Table ~\ref{table:demo} however, shows that certain mental disorders, for example, depression, are closer to females, and drug-related mental disorders are closer to males. This is most likely because most of the males are diagnosed with drug-related disorders and females with depression, bipolar, or anxiety-related disorders. The aforementioned diagnosis ratios with the addition of pre-training create gender-specific biases without that gender being the most prevalent diagnosed in that category. 

Although it is expected to have some degree of bias for some medical terms, such as 'anxiety' which is correlated with gender, we also observed some strong biases that do not exist in or are controversial with medical literature. For example; \textit{Bipolar Disorder, Schizophrenia, and OCD} are biased towards females for both embeddings whereas medicine literature states that there are no marked gender differences in the diagnosis rates of those diseases. Consequently, this ill-founded relationship might cause undesired outcomes in the downstream tasks. In future work, we would like to analyze the effect of those incorrect relationships existing in clinical embeddings to downstream models. 
Moreover, we provide an exhaustive analysis of the demographics of the MIMIC-III dataset in this study. However, we only analyze the gender bias in the embeddings. Bias analysis on different sensitive groups such as race and intersected groups can be studied in future research. 

Another research question of interest for subsequent research could also consider genetic factors. As mentioned earlier in Section~\ref{sec:descriptive}, embeddings are trained on the MIMIC-III dataset which is obtained from hospitals in Boston, USA. However, mental disorders might be affected by both genetic and cultural factors meaning that disease relation to gender might not generalize very well to other countries. This might cause a diagnosis model -which is trained on those clinical embeddings- that performs less fair and accurate for some cultures and countries. We leave this analysis as a future work of this study. 
 
 \subsection{Ethical Considerations}
 Even though the results have shown contextualized word embeddings to be more effective at reducing bias than their non-contextualized counterpart, there are certain ethical considerations to be discussed regarding the decision to implement machine learning systems in the healthcare domain. Due to the high-stakes nature of the healthcare domain, it could be argued that, even though overall biased is significantly reduced with contextual embeddings, the presence of illnesses with high bias scores alone is enough of a reason not to involve machine learning systems in the diagnosing process. Overrepresentation of one diagnosis leads to the underrepresentation of another, the effect of which could be amplified through a feedback loop. Additionally, using a diagnosing system to identify illnesses with low bias scores exclusively is unfeasible due to the illness being unidentified during the diagnosing process.
 Another ethical concern involves the identification of accurate and conflicting biases. Even though experts and statistics are most likely to give an accurate reflection of which category an identified bias belongs to, research has found a gender bias in educational resources related to healthcare \cite{arslanian2006symptoms} and found that experts unintentionally hold negative stereotypes \cite{deangelis2019does}. These biases translate to real-life cases, which means that training data and statistics based on real-life cases inherently hold bias. The potential influence of bias within experts and statistics should be kept in mind when determining to what extent certain biases are accurate and other biases are conflicting.
 
\section{Acknowledgements}
We would like to thank Heysem Kaya, Dong Nguyen and Yupei Du for their useful feedback which helped us to improve the quality of our paper.
\bibliographystyle{ACM-Reference-Format}
\bibliography{references}

\end{document}